\documentclass[conference]{IEEEtran}
\IEEEoverridecommandlockouts
\usepackage[top=54pt,bottom=54pt,left=54pt,right=54pt]{geometry}
\usepackage[pdftex]{epstopdf}
\usepackage{cite}
\usepackage{amsmath,amssymb,amsfonts}
\usepackage{algorithmic}
\usepackage{graphicx}
\usepackage{textcomp}
\usepackage{multirow}
\usepackage{subcaption}
\usepackage{graphicx}
\usepackage{comment}
\usepackage{tikz}
\usetikzlibrary{intersections, calc, arrows.meta}
\usepackage{here}
\usepackage{bm}
\usepackage{booktabs}

\newcommand{\figref}[1]{Fig.~\ref{#1}}
\newcommand{\equref}[1]{\eqref{#1}}

\usepackage[whole]{bxcjkjatype} 
\usepackage{subcaption}
\usepackage[normalem]{ulem}

\def\eg{\textit{e.g}\onedot} 
\def\ie{\textit{i.e}\onedot}

\newcommand{\myvec}[1]{\bm{#1}}

\def\eg{{\it e.g.}}

\def\ie{{\it i.e.}}

\newcommand{\honda}[1]{\textcolor{red}{#1}}

\def\BibTeX{{\rm B\kern-.05em{\sc i\kern-.025em b}\kern-.08em
    T\kern-.1667em\lower.7ex\hbox{E}\kern-.125emX}}
\begin{document}
\vspace{24pt}
\title{ 
Real-Time Model Predictive Control of Vehicles with Convex-Polygon-Aware Collision Avoidance in Tight Spaces
}

\author{\IEEEauthorblockN{1\textsuperscript{st} Haruki Kojima}
\IEEEauthorblockA{\textit{Department of Mechanical Systems Engineering} \\
\textit{Nagoya University}\\
Nagoya, Japan \\
kojima.haruki.r7@s.mail.nagoya-u.ac.jp}
\and
\IEEEauthorblockN{2\textsuperscript{nd} Kohei Honda}
\IEEEauthorblockA{\textit{Department of Mechanical Systems Engineering} \\
\textit{Nagoya University}\\
Nagoya, Japan \\
honda.kohei.k3@f.mail.nagoya-u.ac.jp}
\and
\IEEEauthorblockN{3\textsuperscript{rd} Hiroyuki Okuda}
\IEEEauthorblockA{\textit{Department of Mechanical Systems Engineering} \\
\textit{Nagoya University}\\
Nagoya, Japan \\
h\_okuda@nuem.nagoya-u.ac.jp}
\and
\IEEEauthorblockN{4\textsuperscript{th} Tatsuya Suzuki}
\IEEEauthorblockA{\textit{Department of Mechanical Systems Engineering} \\
\textit{Nagoya University}\\
Nagoya, Japan \\
t\_suzuki@nuem.nagoya-u.ac.jp}
}

\maketitle

\begin{abstract}

This paper proposes vehicle motion planning methods with obstacle avoidance in tight spaces by incorporating polygonal approximations of both the vehicle and obstacles into a model predictive control (MPC) framework. 
Representing these shapes is crucial for navigation in tight spaces to ensure accurate collision detection. 
However, incorporating polygonal approximations leads to disjunctive OR constraints in the MPC formulation, which require a mixed integer programming and cause significant computational cost. 
To overcome this, we propose two different collision-avoidance constraints that reformulate the disjunctive OR constraints as tractable conjunctive AND constraints: (1) a Support Vector Machine (SVM)-based formulation that recasts collision avoidance as a SVM optimization problem, and (2) a Minimum Signed Distance to Edges (MSDE) formulation that leverages minimum signed-distance metrics. 
We validate both methods through extensive simulations, including tight-space parking scenarios and varied-shape obstacle courses, as well as hardware experiments on an RC-car platform. 
Our results demonstrate that the SVM-based approach achieves superior navigation accuracy in constrained environments; the MSDE approach, by contrast, runs in real time with only a modest reduction in collision-avoidance performance.
\end{abstract}


\section{Introduction}

\begin{figure}[t]
    \begin{minipage}[t]{\linewidth}
      \centering
      \includegraphics[width=0.8\linewidth]{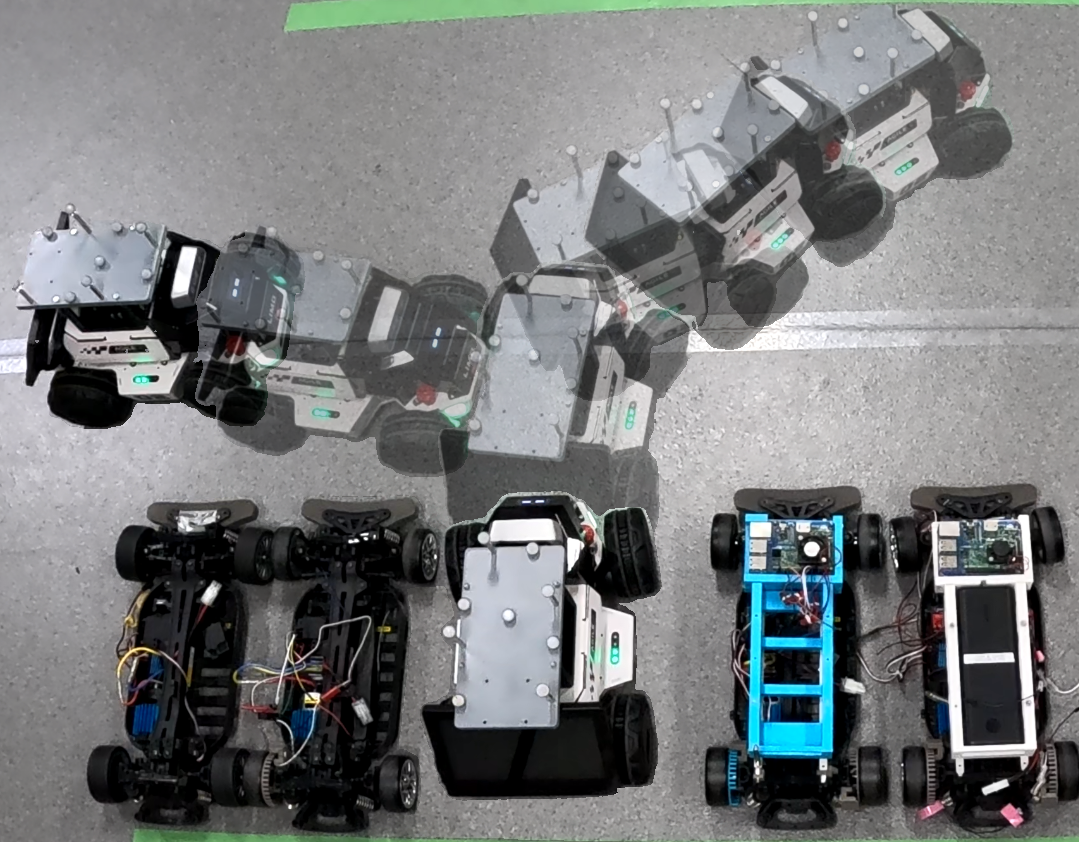}
    \end{minipage}\\
    \begin{minipage}[t]{\linewidth}
      \centering
      \includegraphics[width=0.8\linewidth]{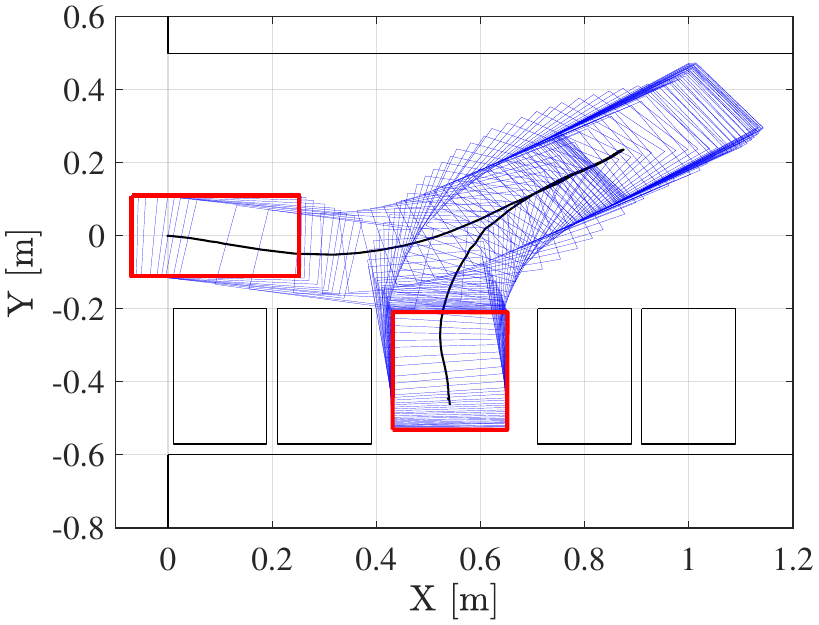}
    \end{minipage}
  \caption{A real-world experiment result using MSDE.}
  \label{fig:experiment_path}
  \vspace{-1em}
\end{figure}

Recent advances in automation have fueled a growing demand for vehicles that can autonomously navigate in extremely tight spaces, \eg, narrow aisles in warehouses or urban parking scenarios. 
In such settings, vehicles often operate with only a few centimeters of clearance from surrounding obstacles. 
Simple reactive controllers, by contrast, struggle in these environments, as they fail to account for both the vehicle's dynamic constraints and the complexity of the workspace.

Model Predictive Control (MPC) provides an attractive framework for tight-space navigation by embedding the vehicle's dynamics and collision-avoidance constraints into an online optimization loop. 
However, existing MPC formulations face a fundamental trade-off between computational efficiency and the accuracy of the collision-avoidance geometry.
Some methods approximate the vehicle and obstacles as circles and ellipses~\cite{circular_approximation1,circular_approximation2,circular_approximation3,circular_approximation4, euclidian_distance, costmap1}, which simplifies the formulation and reduces computational load, but can lead to overly conservative trajectories.
In contrast, approximating with convex polygon models minimizes clearance yet introduces disjunctive \emph{OR} constraints, \ie, separate along edge A or edge B, forcing mixed-integer programs that are too slow for real-time control~\cite{MIP1, MIP2, MIP3}.
Therefore, the central challenge is to express collision avoidance constraints using convex polygons solely with conjunctive \emph{AND} inequalities so that MPC can remain a fast, continuous optimization solvable in real-time.

In this paper, we propose two novel collision avoidance MPC formulations that express convex-polygon constraints as a conjunction of inequalities.
\begin{enumerate}
  \item \emph{Support Vector Machine (SVM)}: Separating-hyperplane formulation for orthogonal convex polygons, which is similar to the prior work of~\cite{hyperplane1}, but we tailored it to the MPC framework and added a regularization term to ensure numerical stability.
  \item \emph{Minimum Signed Distance to Edges (MSDE)}: \emph{OR} to \emph{AND} constraint transformation, which converts disjunctive conditions into an equivalent set of conjunctive inequalities by leveraging the minimum operator.
\end{enumerate}

We validate our approaches through extensive simulations, including tight-space parking scenarios and obstacle courses with varied shapes, as well as hardware experiments on an RC-car platform.
These experiments demonstrate that SVM shows superior success rates and path efficiency in tight spaces compared to MSDE.
However, SVM's computational time is significantly longer than MSDE's, making it impractical for real-time applications.
Thus, we conclude that SVM is effective for offline motion planning, while MSDE is suitable for real-time control in hardware experiments.

\section{Related Work}

\subsection{Classical Obstacle Representations in MPC}
Early mobile-robot MPC schemes rely on voxelised \emph{cost maps}~\cite{costmap1,costmap2}: the accumulated map value along a sampled trajectory penalises collisions but remains non-differentiable, restricting the optimiser to sampling-based methods and requiring high computational resources to ensure approximate optimality~\cite{williams2018information}.

Computationally efficient gradient-based MPC solvers, such as~\cite{Andersson2018, stella2017simple, ohtsuka2004continuation}, require analytic and differentiable formulations of the obstacle avoidance constraints. 
The potential-field method represents the robot and obstacles as a set of soft constraints, which is capable of handling complex shapes, but it does not guarantee collision avoidance and requires careful tuning of the repulsive forces~\cite{APF1, APF2, APF3}.
Approximating bodies by circles and ellipses restores strict separation through a single convex inequality, yet inflates clearance and sacrifices maneuverability in narrow environments~\cite{circular_approximation1,circular_approximation2,circular_approximation3,circular_approximation4, euclidian_distance, costmap1}.

\subsection{Collision Avoidance with Convex-Polygon Representations}

Modeling bodies as convex polygons eliminates artificial clearance but introduces a disjunctive \emph{OR} condition, \ie, the logical disjunction of several linear inequalities, for collision avoidance. This leads to mixed-integer nonlinear programs (MINLPs) that are too slow for real-time control~\cite{MIP1,MIP2,MIP3}. To address this, several methods convert the \emph{OR} condition into a conjunctive \emph{AND} form.

One approach projects a rectangular footprint into a circle on a manifold~\cite{rectangle_to_circle}, enabling tight-space parking at low cost but restricted to rectangular shapes. Other works apply Gale's theorem~\cite{farkas1,farkas2} or the separating hyperplane theorem~\cite{hyperplane1,hyperplane2}, asserting non-intersection if suitable hyperplanes exist. However, these formulations introduce additional decision variables, suffer from numerical divergence, and yield dual-linear constraints whose gradients can be discontinuous, making real-time convergence impractical.

In contrast, our SVM formulation also leverages the separating hyperplane theorem but adds a margin-maximization regularizer to improve convergence of the hyperplane parameters. Our MSDE method requires neither coordinate transformations nor extra variables: its collision constraints are expressed using only linear and \emph{MIN} functions, which support efficient algorithmic differentiation~\cite{algorithmic_differentiation}, resulting in significantly lower computational cost.

\section{Formulation of Collision-Avoidance MPC}

This section formulates the optimal control problem of MPC since this paper proposes convex-polygon-based collision avoidance constraints for MPC.

\subsection{Formation of Vehicle Dynamics}

\begin{figure}[t]
    \centering
    \includegraphics[width=0.65\linewidth]{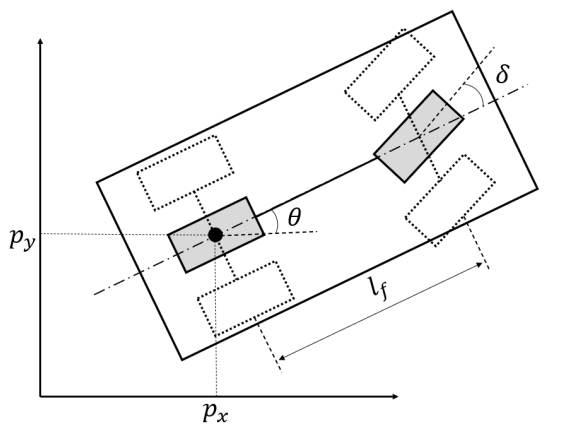}
    \caption{Kinematic Bicycle Model}
    \label{fig:KBM}
    \vspace{-1.0em}
\end{figure}
 
We first define the state and control input vector and model the vehicle dynamics since MPC predicts an infinite sequence of vehicle states to find the optimal control inputs sequence. 
In this paper, the state vectors $\bm{x}$ and the control input vectors $\bm{u}$ for the $k$-th step future at time $t$ are defined as follows:
\begin{align}
  &\bm{x}(k|t)=[p_x(k|t),p_y(k|t),v(k|t),\theta(k|t),\delta(k|t)
  ]^{\top},\\
  &\bm{u}(k|t)=[\dot{v}(k|t),\dot{\delta}(k|t)]^{\top}, 
\end{align}
where $\myvec{p}_{\rm{car}}=[p_x(k|t),p_y(k|t)]^{\top}$ is the position of the vehicle's center of rear axle, 
$v$ is the vehicle's speed, $\theta$ is the yaw angle of the vehicle, and $\delta$ is the tire angle of the vehicle, as shown in Fig.~\ref{fig:KBM}.

We model the vehicle dynamics using the Kinematic Bicycle Model (KBM)~\cite{KBM}, which approximates the vehicle dynamics in the low-speed region. 
The discrete-time state equation of KBM is given as follows:
\begin{align}
  \bm{x}(k+1|t)=\bm{x}(k|t)+
  \begin{bmatrix}
    v(k|t)\cos\theta(k|t) \\ 
    v(k|t)\sin\theta(k|t) \\ 
    \dot{v}(k|t) \\
    \frac{v(k|t)}{l_{\mathrm{f}}}\tan\delta(k|t) \\
    \dot{\delta}(k|t)
  \end{bmatrix} 
  \Delta\tau,
	\label{eq:KBM}
\end{align}
where $l_f$ is the distance from the center of the rear axle to the center of the front axle, and $\Delta\tau$ is the time interval of the predictive horizon.

\subsection{Formulation of Cost Function} \label{sec:evaluate_function}

We then formulate the cost function to reach the goal states $\bm{x}_{\mathrm{ref}}$ smoothly as follows:
\begin{align}
  &J(\bm{x}(k|t),\bm{u}(k|t))=\Phi(\bm{x}(N|t))+\sum_{k=1}^{N-1}L(\bm{x}(k|t),\bm{u}(k|t)), \label{eq:cost_function}\\
	&\Phi(\bm{x}(N|t))=(\bm{x}(N|t)-\bm{x}_{\mathrm{ref}})^{\top}S_{\mathrm{f}}\ (\bm{x}(N|t)-\bm{x}_{\mathrm{ref}}), \\
  &L(\bm{x}(k|t),\bm{u}(k|t))= (\bm{x}(k|t)-\bm{x}_{\mathrm{ref}})^{\top}Q\ (\bm{x}(k|t)-\bm{x}_{\mathrm{ref}})\notag\\
  &\quad+(\bm{u}(k|t)-\bm{u}(k-1|t))^{\top}R\ (\bm{u}(k|t)-\bm{u}(k-1|t)),
\end{align}
where $N$ is the predictive horizon. $\Phi$ and $L$ are the terminal cost and the stage cost, respectively.
$S_{\mathrm{f}}$, $Q$, and $R$ are the weight matrices of the terminal cost and stage cost, respectively.  


\subsection{Formulation of Optimal Control Problem}
\label{sec:optimal_control_problem}

Based on the vehicle dynamics and cost function above, we solve the following optimal control problem at time $t$:
\begin{align}
&\textbf{Given:}
	&&\bm{x}(0|t),\\
&\textbf{Find:}
	&&\bm{x}(k|t),\ k\in\{1,...,N\},\\
	&&&\bm{u}(k|t),\ k\in\{1,...,N-1\},\\
&\textbf{Min.:}
	&& \eqref{eq:cost_function},\notag\\
&\textbf{S.t.:}\
	&& \text{\eqref{eq:KBM}}, \notag\\
  &&& \bm{x}_{\min}\leq\bm{x}(k|t)\leq\bm{x}_{\max},\label{eq:mechanical_state_constraint}\\
  &&& \bm{u}_{\min}\leq\bm{u}(k|t)\leq\bm{u}_{\max},\label{eq:mechanical_input_constraint} \\
	&&& \text{Collision Avoidance Constraints}, \label{eq:collision_avoidance_constraints}
\end{align}
where $\{\bm{x}_{\max},\bm{x}_{\min}\}\in\mathbb{R}^5,\{\bm{u}_{\max},\bm{u}_{\min}\}\in\mathbb{R}^2$ are the upper and lower limits of the vehicle's states and control inputs.
In this work, we propose the collision avoidance constraints \eqref{eq:collision_avoidance_constraints}, which approximate the vehicle's and obstacles' shapes as convex polygons to work in tight spaces.

\section{Convex-polygon-aware\\Collision Avoidance Constraints}

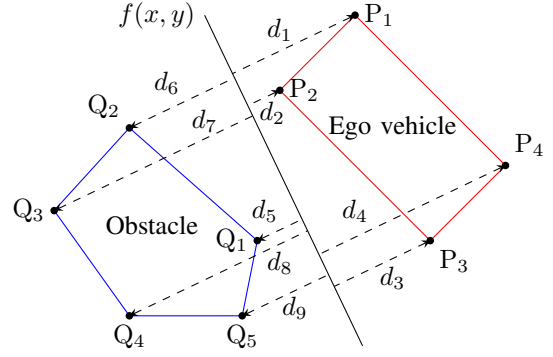
\begin{figure}[t]
  \centering
  \begin{tikzpicture}
    \coordinate[label=right:$\mathrm{P}_1$] (P1) at (1,5);
    \coordinate[label=right:$\mathrm{P}_2$] (P2) at (0,4);
    \coordinate[label=below right:$\mathrm{P}_3$] (P3) at (2,2);
    \coordinate[label=above right:$\mathrm{P}_4$] (P4) at (3,3);
    \draw[red] (P1)--(P2)--(P3)--(P4)--cycle;
    \draw (1.5,3.5)node{Ego vehicle};
    \draw (0,4.8)node{$d_1$};
    \draw (-0.1,3.7)node{$d_2$};
    \draw (1.5,1.5)node{$d_3$};
    \draw (1,2.4)node{$d_4$};
    \coordinate[label=left:$\mathrm{Q}_1$] (Q1) at (-0.3,2);
    \coordinate[label=above left:$\mathrm{Q}_2$] (Q2) at (-2,3.5);
    \coordinate[label=left:$\mathrm{Q}_3$] (Q3) at (-3,2.4);
    \coordinate[label=below:$\mathrm{Q}_4$] (Q4) at (-2,1);
    \coordinate[label=below:$\mathrm{Q}_5$] (Q5) at (-0.5,1);
    \draw[blue] (Q1)--(Q2)--(Q3)--(Q4)--(Q5)--cycle;
    \draw (-1.7,2.2)node{Obstacle};
    \draw (-0.2,2.4)node{$d_5$};
    \draw (-1.5,4.1)node{$d_6$};
    \draw (-1,3.6)node{$d_7$};
    \draw (0,1.7)node{$d_8$};
    \draw (0.2,1.1)node{$d_9$};
    \foreach \P in {P1,P2,P3,P4,Q1,Q2,Q3,Q4,Q5} \fill[black] (\P) circle (0.05);
    \coordinate (R1) at (-1,5);
    \coordinate (R2) at (1.1,0.6);
    \foreach \P in {P1,P2,P3,P4,Q1,Q2,Q3,Q4,Q5} \draw [<-,>=stealth,dashed] (\P)--($(R1)!(\P)!(R2)$);
    \draw (R1)node[left]{$f(x,y)$}--(R2);
  \end{tikzpicture}
  \caption{Support Vector Machine}
  \label{fig:SVM}
\end{figure}

This paper proposes two different collision avoidance constraints for \eqref{eq:collision_avoidance_constraints} by approximating the ego vehicle and obstacles as convex polygons.
We assume that the ego vehicle and obstacle are a convex $n_{\mathrm{ego}}$-gon and $n_{\mathrm{obs}}$-gon, respectively.
The ego vehicle's vertices and obstacle's ones are defined as 
$\mathrm{P}_i(i\in\{1,...,n_{\mathrm{ego}}\})$ and $\mathrm{Q}_i(i\in\{1,...,n_{\mathrm{obs}}\})$, respectively, as shown in Figs.~\ref{fig:SVM} and \ref{fig:MSDE}.
Note that the vertices' positions of the ego vehicle within the predictive horizon can be calculated based on the vehicle's position $\bm{p}_{\mathrm{car}}$ and yaw angle $\theta$ if the vehicle's shape is known.

\subsection{Support Vector Machine-based Constraints}

Our first proposed method is based on the idea of the previous work in \cite{hyperplane1}.
According to the separating hyperplane theorem, two convex shapes do not intersect if there exists a hyperplane separating them. 
Figure~\ref{fig:SVM} illustrates an ego vehicle, obstacle, and separating line. We label the ego vehicle's vertices, $\mathrm{P}_i$ for $i\in \{1,\dots,n_{\mathrm{ego}} \}$, with a positive class and obstacle's vertices, $\mathrm{Q}_j$ for $j\in \{1,\dots,n_{\mathrm{obs}} \}$, with a negative class.
For convenience, we define the concatenated vertices vector $\bm{p}=(\mathrm{P}_1, \dots, \mathrm{P}_{n_{\mathrm{ego}}}, \mathrm{Q}_1, \dots, \mathrm{Q}_{n_{\mathrm{obs}}} )=\{\bm{p}_k\}
(k \in \{1, \dots, n_{\mathrm{ego}}, n_{\mathrm{ego}}+1, n_{\mathrm{ego}}+n_{\mathrm{obs}}\}$) 
with class labels $q_k=-1(k\in \{ 1, \dots, n_{\mathrm{ego}} \})$ and $q_k=+1(k\in \{ n_{\mathrm{ego}}+1, \dots, n_{\mathrm{ego}}+n_{\mathrm{obs}}\})$.
If a line $f(x,y)=ax+by+c=0$ that separates these two sets of vertices is found, the ego vehicle does not collide with the obstacle.
The signed distance margin $d_k$ between the vertex ${\bm{p}}_k=(x_k,y_k)$ and the separating line $f(x,y)$ is calculated as
\begin{align}
  d_k=\frac{q_k f(\bm{p}_k)}{\sqrt{a^2+b^2}}=\frac{q_k(ax_k+by_k+c)}{\sqrt{a^2+b^2}}, \forall k 
	\label{eq:margin}
\end{align}
and requiring 
\begin{align}
  \forall k,\quad q_k(ax_k+by_k+c)=q_kf(x_k,y_k)>0 \label{eq:hyperplane_constraint},
\end{align}
to ensure that all margins are positive, implying the two convex polygons are separable and do not intersect.

The previous work~\cite{hyperplane1} uses \eqref{eq:hyperplane_constraint} directly as a collision-avoidance constraint.
However, because the separating line is not unique, its parameters $(a,b,c)$ may diverge. 
To address this, we adopt the SVM formulation to compute a canonical separating line.
In the SVM formulation, we maximize the minimum margin:
\begin{align}
	\max_{a,b,c\in\mathbb{R}}
	\left(\min_{k\in\{1,...,n_{\mathrm{ego}}+n_{\mathrm{obs}}\}}\left(\frac{q_k(ax_k+by_k+c)}{\sqrt{a^2+b^2}}\right)\right).
    \label{eq:original_svm}
\end{align}
Let vertex $\bm{p}_l(x_l,y_l)$ attain this minimum margin. 
Then \eqref{eq:original_svm} is equivalent to 
\begin{align}
  &\max_{a,b,c\in\mathbb{R}}\frac{q_l(ax_l+by_l+c)}{\sqrt{a^2+b^2}}\\
  &\mathrm{s.t.}\quad q_k(ax_k+by_k+c)\geq q_l(ax_l+by_l+c)\quad \forall k \label{eq:min_constraint}.
\end{align}
In \eqref{eq:min_constraint}, by multiplying the parameters $a,b,c$ by a constant such that $q_l(ax_l+by_l+c)=\epsilon>0$ is satisfied, the problem above becomes equivalent to the following problem:
\begin{align}
  &\max_{a,b,c\in\mathbb{R}}\frac{\epsilon}{\sqrt{a^2+b^2}}, \label{eq:maximize_margin}\\
  &\mathrm{s.t.}\quad q_k(ax_k+by_k+c)\geq\epsilon\quad\forall k,
  \label{eq:line_constraint}
\end{align}
where we set $\epsilon=10^{-6}$. 
This allows the parameters of the separating line to exist in a wider range, making them easier to find.
When the evaluation function in \equref{eq:maximize_margin} is squared and taken as its reciprocal, the problem above becomes approximately equivalent to the following problem:
\begin{align}
  &\min_{a,b,c\in\mathbb{R}}\alpha(a^2+b^2), \label{eq:svm_regulation} \\
  &\mathrm{s.t.}\quad q_k(ax_k+by_k+c)\geq\epsilon\quad\forall k. \label{eq:svm_constraint}
\end{align}
\eqref{eq:svm_constraint} is nearly equivalent to \eqref{eq:hyperplane_constraint}, which represents the collision avoidance constraint.
If the parameters $a,b,c$ satisfying the constraint exist, the two convex polygons do not intersect, \ie, the ego vehicle does not collide with the obstacle.

We incorporate the above SVM formulation into the optimal control problem described in Sec.~\ref{sec:optimal_control_problem} as a collision avoidance constraint.
The parameters $a,b,c$ are added to the decision variables, \eqref{eq:svm_constraint} is added to the collision avoidance constraints in \eqref{eq:collision_avoidance_constraints},
and \eqref{eq:svm_regulation} is added to the cost function in \eqref{eq:cost_function} as a regularization term.
Note that the regularization term is multiplied by a small coefficient $\alpha=10^{-4}$ to ensure that the optimization of vehicle control takes precedence over margin maximization.
As a result, the SVM-based collision avoidance constraints are \emph{AND} conditions, and their number equals the sum of the vertices of the ego vehicle and obstacle.

\subsection{Minimum Signed Distance to Edges-based Constraints}

\begin{figure}[t]
  \begin{minipage}[t]{0.45\hsize}
  \centering
  \begin{tikzpicture}
    \coordinate[label=above left:$\mathrm{P}_1$] (P1) at (-1.5,3.914);
    \coordinate[label=above left:$\mathrm{P}_2$] (P2) at (-1.5,2.5);
    \coordinate[label=below right:$\mathrm{P}_3$] (P3) at (1.328,2.5);
    \coordinate[label=above right:$\mathrm{P}_4$] (P4) at (1.328,3.914);
    \draw[red] (P1)--(P2)--(P3)--(P4)--cycle;
    \draw (0,3.5)node{Ego vehicle};
    \coordinate[label=right:$\mathrm{Q}_1$] (Q1) at (-0.3,2);
    \coordinate[label=left:$\mathrm{Q}_2$] (Q2) at (-2,3.5);
    \coordinate[label=left:$\mathrm{Q}_3$] (Q3) at (-3,2.4);
    \coordinate[label=below left:$\mathrm{Q}_4$] (Q4) at (-2,1);
    \coordinate[label=below right:$\mathrm{Q}_5$] (Q5) at (-0.5,1);
    \draw[blue] (Q1)--(Q2)--(Q3)--(Q4)--(Q5)--cycle;
    \draw (-1.4,1.6)node{Obstacle};
    \coordinate[label=left:G] (G) at (-2,2.2);
    \draw (-1.2,2.3)node{$d_2$};
    \draw (-0.8,2.8)node{$l_1$};
    \draw (-2.8,3)node{$l_2$};
    \draw (-2.8,1.8)node{$l_3$};
    \draw (-1.3,0.7)node{$l_4$};
    \draw (-0.1,1.5)node{$l_5$};
    \draw (2,1)node{$d_2\ge0$};
    \foreach \P in {P1,P2,P3,P4,Q1,Q2,Q3,Q4,Q5,G} \fill[black] (\P) circle (0.05);
    \draw[dashed] ($(Q1)!-0.8!(Q2)$)--(Q1);
    \draw[dashed] ($(Q2)!-0.8!(Q3)$)--(Q2);
    \draw[dashed] ($(Q3)!-0.8!(Q4)$)--(Q3);
    \draw[dashed] ($(Q4)!-0.8!(Q5)$)--(Q4);
    \draw[dashed] ($(Q5)!-0.8!(Q1)$)--(Q5);
    \draw[dashed] (Q2)--($(Q1)!1.6!(Q2)$);
    \draw[dashed] (Q3)--($(Q2)!1.6!(Q3)$);
    \draw[dashed] (Q4)--($(Q3)!1.6!(Q4)$);
    \draw[dashed] (Q5)--($(Q4)!1.6!(Q5)$);
    \draw[dashed] (Q1)--($(Q5)!1.6!(Q1)$);
    \draw[->,>=stealth] ($(Q1)!(P2)!(Q2)$)--(P2);
    \draw[dotted] (-4,0)--(4,0);
  \end{tikzpicture}
\end{minipage} \\\\
\begin{minipage}[t]{0.45\hsize}
  \centering
  \begin{tikzpicture}
    \coordinate[label=above left:$\mathrm{P}_1$] (P1) at (0.5,3.914);
    \coordinate[label=above right:$\mathrm{P}_2$] (P2) at (0.5,2.5);
    \coordinate[label=below right:$\mathrm{P}_3$] (P3) at (3.328,2.5);
    \coordinate[label=above right:$\mathrm{P}_4$] (P4) at (3.328,3.914);
    \draw[red] (P1)--(P2)--(P3)--(P4)--cycle;
    \draw (2,3.5)node{Ego vehicle};
    \coordinate[label=right:$\mathrm{Q}_1$] (Q1) at (-0.3,2);
    \coordinate[label=left:$\mathrm{Q}_2$] (Q2) at (-2,3.5);
    \coordinate[label=left:$\mathrm{Q}_3$] (Q3) at (-3,2.4);
    \coordinate[label=below left:$\mathrm{Q}_4$] (Q4) at (-2,1);
    \coordinate[label=below right:$\mathrm{Q}_5$] (Q5) at (-0.5,1);
    \draw[blue] (Q1)--(Q2)--(Q3)--(Q4)--(Q5)--cycle;
    \draw (-1.4,1.6)node{Obstacle};
    \coordinate[label=left:G] (G) at (-2,2.2);
    \draw (0.2,2.8)node{$d_2$};
    \draw (-0.8,2.8)node{$l_1$};
    \draw (-2.8,3)node{$l_2$};
    \draw (-2.8,1.8)node{$l_3$};
    \draw (-1.3,0.7)node{$l_4$};
    \draw (-0.1,1.5)node{$l_5$};
    \draw (2,1)node{$d_2<0$};
    \foreach \P in {P1,P2,P3,P4,Q1,Q2,Q3,Q4,Q5,G} \fill[black] (\P) circle (0.05);
    \draw[dashed] ($(Q1)!-0.8!(Q2)$)--(Q1);
    \draw[dashed] ($(Q2)!-0.8!(Q3)$)--(Q2);
    \draw[dashed] ($(Q3)!-0.8!(Q4)$)--(Q3);
    \draw[dashed] ($(Q4)!-0.8!(Q5)$)--(Q4);
    \draw[dashed] ($(Q5)!-0.8!(Q1)$)--(Q5);
    \draw[dashed] (Q2)--($(Q1)!1.6!(Q2)$);
    \draw[dashed] (Q3)--($(Q2)!1.6!(Q3)$);
    \draw[dashed] (Q4)--($(Q3)!1.6!(Q4)$);
    \draw[dashed] (Q5)--($(Q4)!1.6!(Q5)$);
    \draw[dashed] (Q1)--($(Q5)!1.6!(Q1)$);
    \draw[<-,>=stealth] ($(Q1)!(P2)!(Q5)$)--(P2);
  \end{tikzpicture}
\end{minipage}
  \caption{Collision avoidance based on minimum signed distance. The collision between the ego vehicle and the obstacle can be detected by checking whether the ego vehicle's vertex $\mathrm{P}_2$, which has the minimum signed distance $d_2$, is inside (top) or outside (bottom) the obstacle.}
  \label{fig:MSDE}
\end{figure}
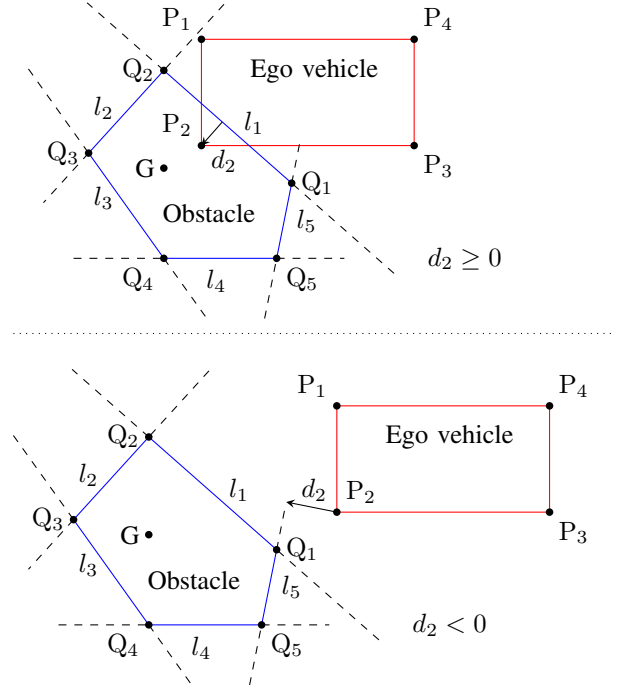

The second proposed method is based on the concept of signed distances to edges. 
We detect collisions between two convex polygons by checking whether every vertex of one polygon lies inside the other, as shown in \figref{fig:MSDE}.
An obstacle edge $i$ can be represented by the line $l_i(x,y)=\alpha_i x + \beta_i y + \gamma_i = 0$
passing through vertices $\mathrm{Q}_i$ and $\mathrm{Q}_{i+1}$.  If a point $\mathrm{G}(x_{\mathrm{G}},y_{\mathrm{G}})$ lies inside the obstacle, there must exist at least one such line satisfying
\begin{align}
  \exists\, i\in\{1,\dots,n_{\mathrm{obs}}\}, \;
  l_i(x_{\mathrm{G}},y_{\mathrm{G}})=\alpha_i x_{\mathrm{G}} + \beta_i y_{\mathrm{G}} + \gamma_i \ge 0.
\end{align}
When an ego vehicle's vertex $\mathrm{P}_j(x_j,y_j)$ intrudes into the obstacle, it must satisfy the following \emph{AND} condition:
\begin{align}
  \forall\, i\in\{1,\dots,n_{\mathrm{obs}}\}, \;
  l_i(x_j,y_j)=\alpha_i x_j + \beta_i y_j + \gamma_i \ge 0,
\end{align}
which means that $\mathrm{P}_j$ lies on the same side of every obstacle edge as the interior point $\mathrm{G}$.

Conversely, avoiding collision requires that each vertex violate at least one of these inequalities, giving the following \emph{OR} condition:
\begin{align}
  \exists\, i\in\{1,\dots,n_{\mathrm{obs}}\}, \;
  l_i(x_j,y_j)=\alpha_i x_j + \beta_i y_j + \gamma_i < 0.
  \label{eq:point_avoidance1}
\end{align}
However, embedding such \emph{OR} constraints in an optimal control problem leads to a mixed-integer program with binary variables, incurring high computational cost in time and memory~\cite{MIP1, MIP2, MIP3}. 
To avoid this, we introduce a minimum operator and rewrite \equref{eq:point_avoidance1} as
\begin{align}
  d_j = \min_i \bigl(l_i(x_j,y_j)\bigr)
      = \min_i \bigl(\alpha_i x_j + \beta_i y_j + \gamma_i\bigr) < 0,
  \label{eq:EL_constraint}
\end{align}
where $d_j$ is the minimum signed distance from all obstacle-edge lines to vertex $\mathrm{P}_j$. As $\mathrm{P}_j$ approaches the obstacle, $d_j$ increases; if $d_j \ge 0$, then $\mathrm{P}_j$ lies inside the obstacle, as shown in \figref{fig:MSDE}.

Finally, by enforcing \equref{eq:EL_constraint} for every vertex of both the ego vehicle and obstacle, we obtain a complete set of collision-avoidance constraints to incorporate into \eqref{eq:collision_avoidance_constraints}.
As well as the SVM-based method, the MSDE-based collision avoidance constraints consist of only AND conditions and the number of constraints equals the sum of the vertex counts of the ego vehicle and obstacle.

\subsection{Application to Circular Obstacles}

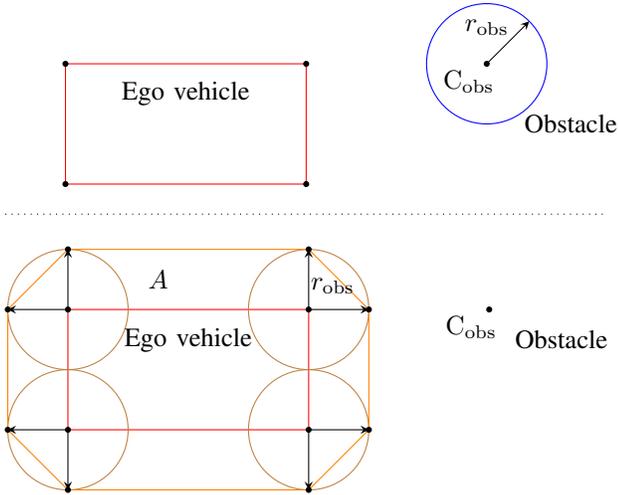
\begin{figure}[t]
  \begin{minipage}[t]{0.45\hsize}
  \centering
    \begin{tikzpicture}[scale=0.8]
    \coordinate (P1) at (1,2);
    \coordinate (P2) at (1,0);
    \coordinate (P3) at (5,0);
    \coordinate (P4) at (5,2);
    \draw[red] (P1)--(P2)--(P3)--(P4)--cycle;
    \coordinate (C) at (8,2);
    \draw (7.7,1.7)node{$\mathrm{C_{obs}}$};
    \draw[->,>=stealth] (C)--(8.707,2.707);
    \draw (8,2.6)node{$r_{\mathrm{obs}}$};
    \draw (9.4,1)node{Obstacle};
    \draw[blue] (C) circle (1);
    \foreach \P in {P1,P2,P3,P4,C} \fill[black] (\P) circle (0.05);
    \draw (3,1.5)node{Ego vehicle};
    \draw[dotted] (0,-0.5)--(10,-0.5);
  \end{tikzpicture}
  \end{minipage}\\\\
  \begin{minipage}[t]{0.45\hsize}
  \centering
    \begin{tikzpicture}[scale=0.8]
    \coordinate (P1) at (1,2);
    \coordinate (P2) at (1,0);
    \coordinate (P3) at (5,0);
    \coordinate (P4) at (5,2);
    \draw[red] (P1)--(P2)--(P3)--(P4)--cycle;
    \foreach \P in {P1,P2,P3,P4} \draw[brown] (\P) circle (1);
    \coordinate (R1) at (1,3);
    \coordinate (R2) at (0,2);
    \foreach \P in {R1,R2} \draw[<-,>=stealth] (\P)--(P1);
    \coordinate (R3) at (0,0);
    \coordinate (R4) at (1,-1);
    \foreach \P in {R3,R4} \draw[<-,>=stealth] (\P)--(P2);
    \coordinate (R5) at (5,-1);
    \coordinate (R6) at (6,0);
    \foreach \P in {R5,R6} \draw[<-,>=stealth] (\P)--(P3);
    \coordinate (R7) at (6,2);
    \coordinate (R8) at (5,3);
    \foreach \P in {R7,R8} \draw[<-,>=stealth] (\P)--(P4);
    \draw[orange] (R1)--(R2)--(R3)--(R4)--(R5)--(R6)--(R7)--(R8)--cycle;
    \draw (5.4,2.4)node{$r_{\mathrm{obs}}$};
    \coordinate (C) at (8,2);
    \draw (7.7,1.7)node{$\mathrm{C_{obs}}$};
    \draw (9.2,1.5)node{Obstacle};
    \foreach \P in {P1,P2,P3,P4,R1,R2,R3,R4,R5,R6,R7,R8,C} \fill[black] (\P) circle (0.05);
    \draw (3,1.5)node{Ego vehicle};
    \draw (2.5,2.5)node{$A$};
  \end{tikzpicture}
  \end{minipage}
  \caption{The application of the circular obstacle avoidance. We transform the polygon-to-circle avoidance problem (top) into the polygon-to-polygon avoidance problem (bottom).}
  \label{fig:car_circle}
\end{figure}

We propose two polygon-to-polygon collision avoidance constraints, \ie, SVM and MSDE, in the previous sections.
We lastly describe how to apply these methods to circular obstacles, which are often encountered in real-world scenarios.
Figure \ref{fig:car_circle} shows a situation where the ego vehicle avoids a circular obstacle $\mathrm{C_{obs}}$ with radius $r_{\mathrm{obs}}$.

In both the SVM and MSDE methods, we define a convex region \(A\) by offsetting the ego vehicle's geometry outward by the obstacle's radius \(r_{\mathrm{obs}}\). 
This region \(A\) consists of four brown circles and an orange octagon. Each circle is centered at one of the ego vehicle's vertices and has radius \(r_{\mathrm{obs}}\). The octagon's vertices are obtained by extending the vehicle's bounding rectangle by \(r_{\mathrm{obs}}\) in both the longitudinal and lateral directions.

The collision-avoidance constraint requires that the obstacle center \(\mathrm{C_{obs}}(x_{\mathrm{obs}},y_{\mathrm{obs}})\) lie outside region \(A\). While the constraint for the octagon is given above, the avoidance constraint with each brown circle is:
\begin{align}
\forall i\in\{1,\dots,n_{\mathrm{ego}}\},\quad (x_i - x_{\mathrm{obs}})^2 + (y_i - y_{\mathrm{obs}})^2 > r_{\mathrm{obs}}^2.
\label{eq:circle_avoidance}
\end{align}
\section{Experiments}

\subsection{Simulation Settings}

In our simulations, we compare the proposed avoidance methods in terms of computational cost and navigational performance, \ie, the success rate and completion time.
We evaluate three scenarios: reverse parking, parallel parking, and obstacle avoidance in tight spaces with various obstacle shapes.
Following Japanese standard vehicle dimensions, we set the vehicle's length $l_{\mathrm{car}}=4.0\ \mathrm{m}$, width $w_{\mathrm{car}}=1.7\ \mathrm{m}$, wheelbase $l_{\mathrm{f}}=2.5\ \mathrm{m}$, rear overhang $l_{\mathrm{roh}}=0.8\ \mathrm{m}$, and safety margin $d_{\mathrm{margin}}=0.05\ \mathrm{m}$. 
The vehicle's input limits are: velocity $v_{\mathrm{lim}}=2\ \mathrm{m/s}$, steering angle $\delta_{\mathrm{lim}}=0.70\ \mathrm{rad}$, acceleration $\dot{v}_{\mathrm{lim}}=1\ \mathrm{m/s^2}$, and steering angle rate $\dot{\delta}_{\mathrm{lim}}=6.28\ \mathrm{rad/s}$. Other state variables are unconstrained. All simulations were performed on a laptop equipped with an Intel Core i7-1165G7 CPU (2.80\,GHz, 4 cores) and 16\,GB of RAM.

\begin{figure}[t]
  \begin{minipage}[t]{\linewidth}
    \centering
    \includegraphics[width=0.8\linewidth]{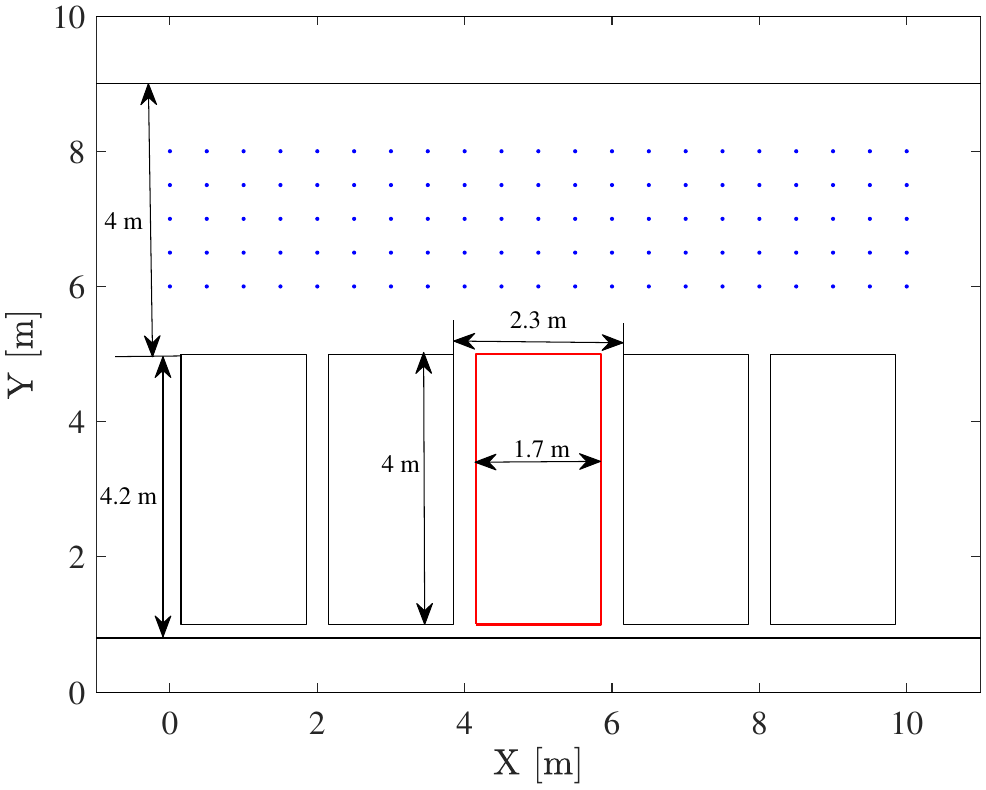}
  \end{minipage}\\
  \begin{minipage}[t]{\linewidth}
    \centering
    \includegraphics[width=0.8\linewidth]{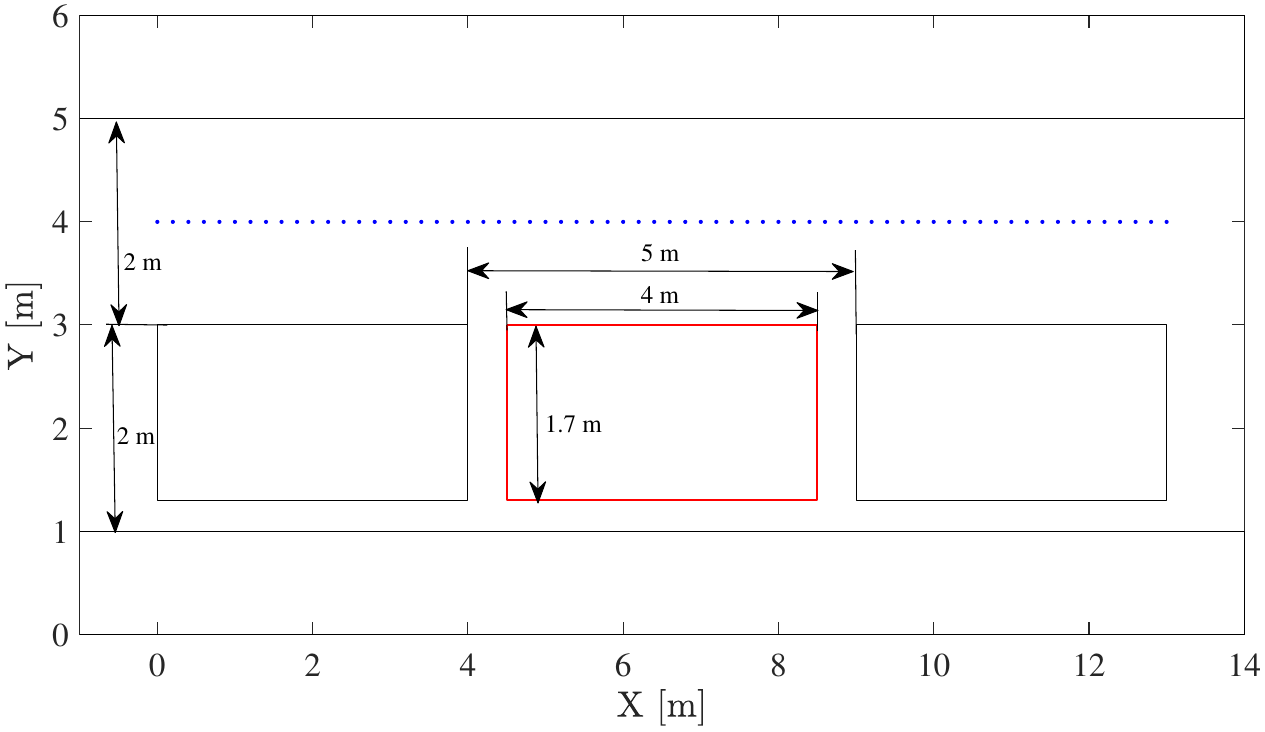}
  \end{minipage}
  \caption{Environments of reverse parking (top) and parallel parking (bottom).}
  \label{fig:environment}
\end{figure}

Figure~\ref{fig:environment} illustrates the environments for the reverse and parallel parking tasks. In each task, the ego vehicle must park among stationary vehicles. Initial positions are sampled uniformly: 105 blue points spaced by $0.5\ \mathrm{m}$ for reverse parking, and 66 points spaced by $0.2\ \mathrm{m}$ for parallel parking. All initial state variables, except for the position, are set to zero. The goal is to park in the designated red parking spots.

\subsection{Implementation Details}

In simulations, the prediction horizon interval $\Delta\tau$ and cycle time $\Delta t$ are fixed at 0.2\,s.
\begin{table}[t]
  \centering
  \small
  \caption{Weight Parameters in the Cost Function}
  \label{tab:parameter}
  \begin{tabular}{@{}lcc@{}}
    \toprule
    & \textbf{Reverse Parking} & \textbf{Parallel Parking} \\
    \midrule
    $S_{\mathrm{f}}$ & $\mathrm{diag}\bigl[300,\,300,\,15,\,600,\,15\bigr]$ & $\mathrm{diag}\bigl[800,\,800,\,20,\,400,\,20\bigr]$ \\
    $Q$              & $\mathrm{diag}\bigl[0.25,\,0.25,\,0.05,\,1,\,0.05\bigr]$ & $\mathrm{diag}\bigl[0.5,\,0.5,\,0.05,\,0.5,\,0.05\bigr]$ \\
    $R$              & $\mathrm{diag}\bigl[0.2,\,20\bigr]$                       & $\mathrm{diag}\bigl[0.2,\,20\bigr]$                       \\
    \bottomrule
  \end{tabular}
\end{table}
Table~\ref{tab:parameter} lists the cost function weights. We assign larger terminal cost weights $S_{\mathrm{f}}$ to penalize deviations between the terminal and reference states. Conversely, the stage cost weights $Q$ for state errors are reduced to allow greater state deviation for safer obstacle avoidance. The input cost weight for acceleration, $r^{\dot{v}}$, is increased to penalize abrupt acceleration changes, whereas the weight for steering rate, $r^{\dot{\delta}}$, is decreased to permit rapid steering adjustments in tight spaces. These weights can be tuned according to each task's requirements.

We use IPOPT~\cite{IPOPT}, which employs a primal-dual interior-point method to solve the resulting nonlinear programming problem efficiently, even with many inequality constraints. 

\subsection{Evaluation Metrics}

We evaluate computational cost by measuring the average and worst-case solving times per MPC iteration. To assess navigation performance, we adopt the success weighted by completion time (SCT) metric~\cite{SCT}:

\begin{align}
  \mathrm{SCT}=\frac{1}{N}\sum_{i=1}^N S_i\frac{T_i}{\max(C_i,T_i)} \label{eq:SCT}, 
\end{align}
where $N$ is the total number of episodes, $S_i\in\{0,1\}$ indicates success of episode $i$, $C_i$ is the completion time for the episode, and $T_i$ is the minimum completion time among all methods. SCT increases with both higher success rates and shorter completion times. An episode is considered successful if the vehicle's position is within $0.2\ \mathrm{m}$ of the reference point and its orientation is within $\pm10^\circ$ of the reference angle.

\subsection{Simulation Results}

\begin{figure}[t]
  \begin{tabular}{cc}
    \begin{minipage}[t]{0.45\hsize}
      \centering
      \includegraphics[width=\linewidth]{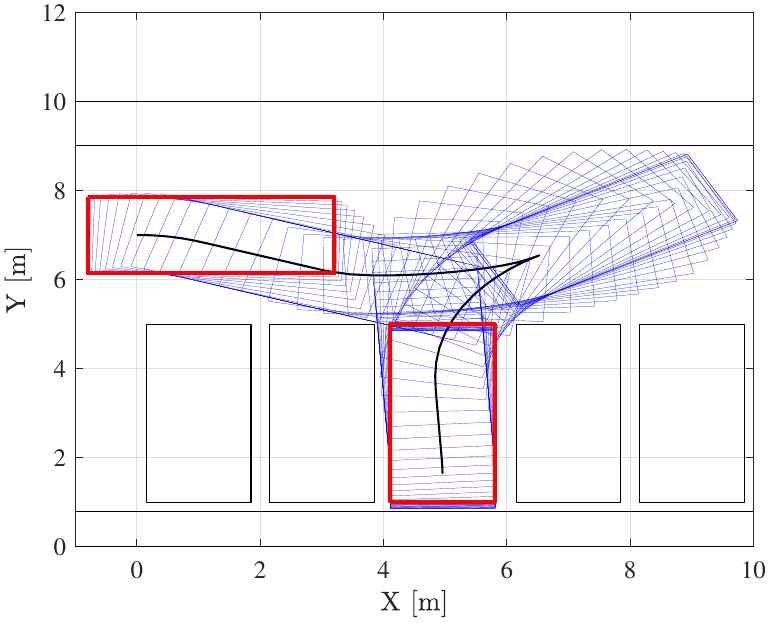}
    \end{minipage}&
    \begin{minipage}[t]{0.45\hsize}
      \centering
      \includegraphics[width=\linewidth]{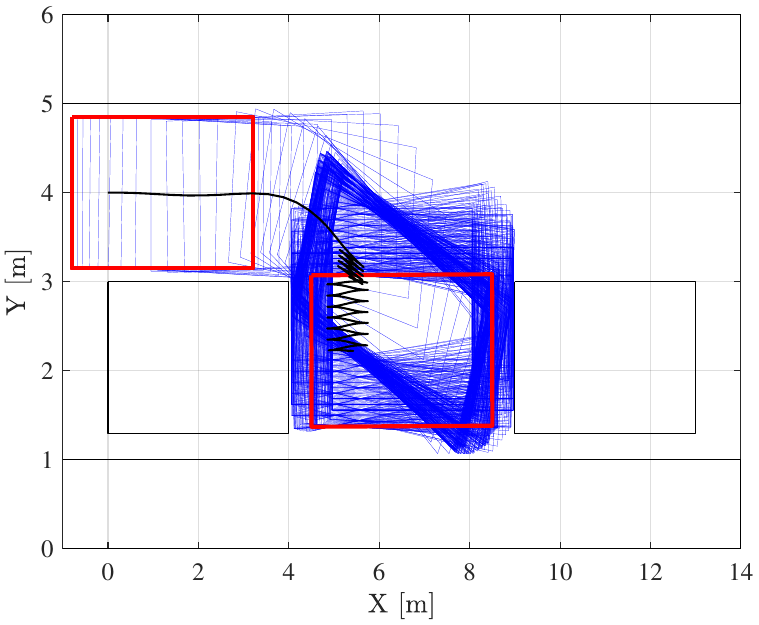}
    \end{minipage}\\
    \begin{minipage}[t]{0.45\hsize}
      \centering
      \includegraphics[width=\linewidth]{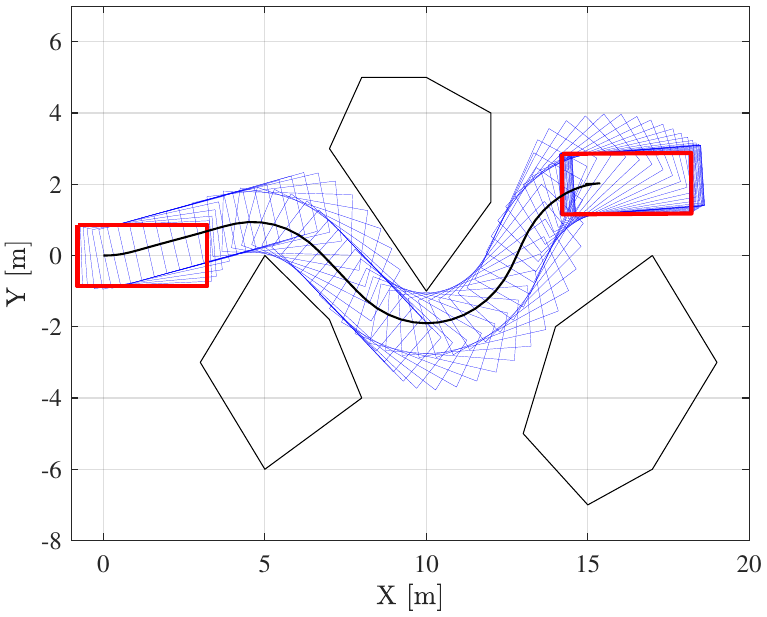}
    \end{minipage}&
    \begin{minipage}[t]{0.45\hsize}
      \centering
      \includegraphics[width=\linewidth]{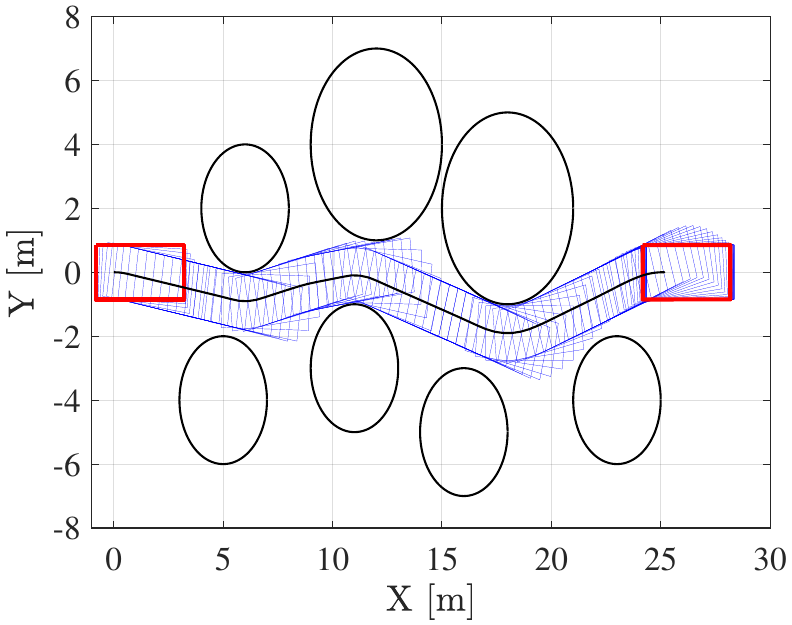}
    \end{minipage}
  \end{tabular}
  \caption{Trajectories with MSDE in reverse parking (upper left), parallel parking (upper right), avoidance of polygons (lower left), and circles (lower right).}
  \label{fig:trajectory}
\end{figure}

Figure~\ref{fig:trajectory} illustrates example trajectories for the MSDE method in reverse parking, parallel parking, polygon avoidance, and circle avoidance. Both SVM and MSDE navigate the parking tasks successfully, maneuvering iteratively without collision.

\begin{figure}[t]
  \begin{tabular}{cc}
    \begin{minipage}[t]{0.45\hsize}
      \centering
      \includegraphics[width=\linewidth]{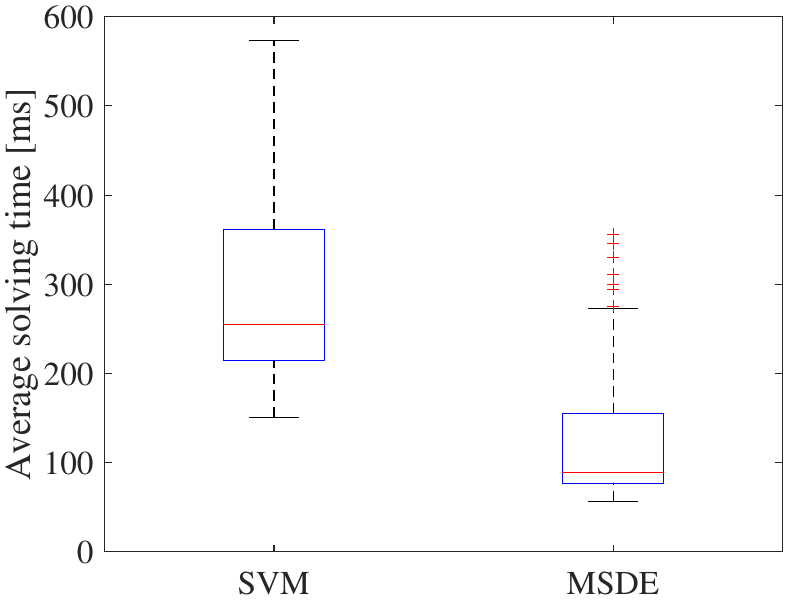}
    \end{minipage}&
    \begin{minipage}[t]{0.45\hsize}
      \centering
      \includegraphics[width=\linewidth]{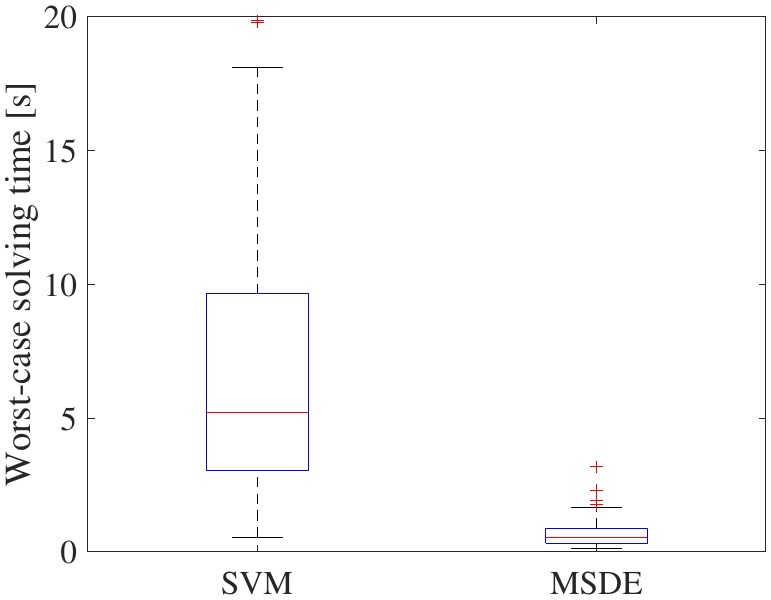}
    \end{minipage}\\
    \begin{minipage}[t]{0.45\hsize}
      \centering
      \includegraphics[width=\linewidth]{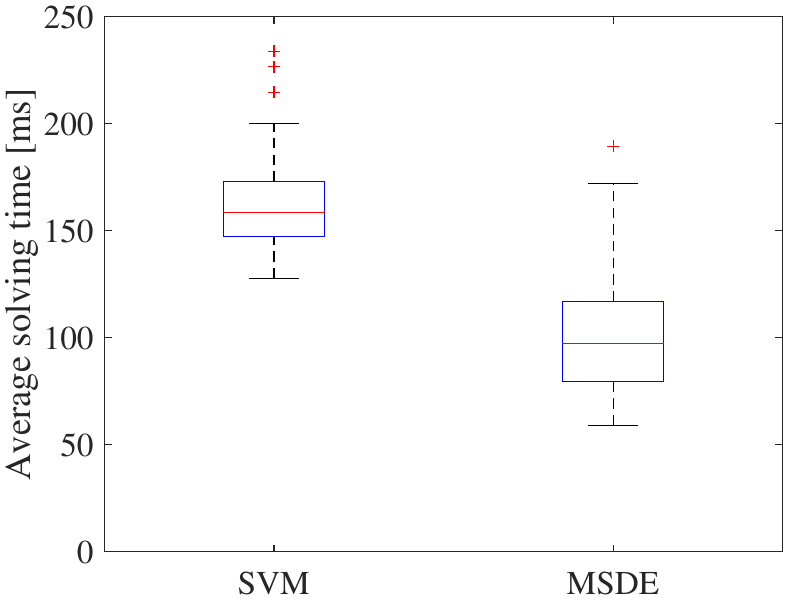}
    \end{minipage}&
    \begin{minipage}[t]{0.45\hsize}
      \centering
      \includegraphics[width=\linewidth]{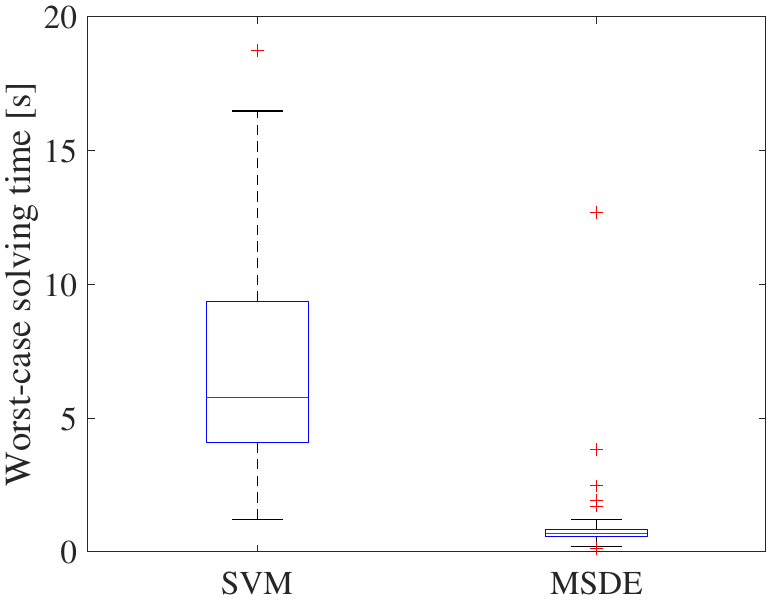}
    \end{minipage}
  \end{tabular}
  \caption{Average and worst-case solving time in reverse parking (upper row) and parallel parking (lower row).}
  \label{fig:solving_time}
\end{figure}

Figure~\ref{fig:solving_time} presents box plots of the average and worst-case solving times for reverse and parallel parking. MSDE consistently achieves lower solving times than SVM. This difference arises because SVM's collision constraint $q_k(ax_k+by_k+c)$ is bilinear, requiring simultaneous optimization over two variables, whereas MSDE's constraint $d_j$ is effectively linear for static obstacles when using algorithmic differentiation of the minimum function~\cite{algorithmic_differentiation}. Consequently, MSDE uses fewer decision variables and simpler gradient computations.

\begin{table}[t]
  \centering
  \small
  \caption{Simulation Results in the Parking Scenarios}
  \label{tab:parking_sct}
  \begin{tabular}{@{}lcc@{\quad}cc@{}}
    \toprule
    \multirow{2}{*}{\textbf{Parking type}} 
      & \multicolumn{2}{c}{\textbf{Reverse}} 
      & \multicolumn{2}{c}{\textbf{Parallel}} \\
    \cmidrule(lr){2-3} \cmidrule(lr){4-5}
      & \textbf{SVM} & \textbf{MSDE} 
      & \textbf{SVM} & \textbf{MSDE} \\
    \midrule
    Number of variables 
      & 523 & 145 
      & 397 & 145 \\
    SCT 
      & \textbf{0.78} & 0.68 
      & \textbf{0.95} & 0.47 \\
    \bottomrule
  \end{tabular}
\end{table}

Table~\ref{tab:parking_sct} compares SCT and the number of decision variables for both methods. 
Although SVM attains a higher SCT than MSDE, due to its larger search space and continuous gradient, it incurs a higher computational cost.

In the obstacle avoidance scenarios (polygons and circles), both methods achieve collision-free navigation with identical completion times. Table~\ref{tab:various_avoid_sct} reports the number of variables, average solving time, and worst-case solving time. As in the parking tasks, MSDE significantly outperforms SVM in computational efficiency.

\begin{table}[t]
  \centering
  \small
  \caption{Computational Times in the Obstacle Avoidance Scenario}
  \label{tab:various_avoid_sct}
  \begin{tabular}{@{}lcc@{\quad}cc@{}}
    \toprule
    \multirow{2}{*}{\textbf{Obstacle type}}
      & \multicolumn{2}{c}{\textbf{Polygons}}
      & \multicolumn{2}{c}{\textbf{Circles}} \\
    \cmidrule(lr){2-3} \cmidrule(lr){4-5}
      & \textbf{SVM} & \textbf{MSDE}
      & \textbf{SVM} & \textbf{MSDE} \\
    \midrule
    Number of variables
      & 271  & 145
      & 586  & 145 \\
    Average solving time [ms]
      & 100  & \textbf{54.0}
      & 622  & \textbf{123} \\
    Worst-case solving time [ms]
      & 511  & \textbf{224}
      & 17901 & \textbf{650} \\
    \bottomrule
  \end{tabular}
\end{table}

\subsection{Real World Experiment Settings}

To validate the methods in a real-world setting, we used a \textit{LIMO Pro}, a ROS-programmable small steering vehicle~\footnote{https://global.agilex.ai/products/limo-pro}.
Its dimensions are set to $l_{\mathrm{car}}=322\ \mathrm{mm}$, $w_{\mathrm{car}}=220\ \mathrm{mm}$, $l_{\mathrm{f}}=200\ \mathrm{mm}$, and $l_{\mathrm{roh}}=70\ \mathrm{mm}$, with a safety margin $d_{\mathrm{margin}}=20\ \mathrm{mm}$. 
We limit its velocity to $v_{\mathrm{lim}}=2\ \mathrm{m/s}$, steering angle to $\delta_{\mathrm{lim}}=0.35\ \mathrm{rad}$, acceleration to $\dot{v}_{\mathrm{lim}}=1\ \mathrm{m/s^2}$, and steering rate to $\dot{\delta}_{\mathrm{lim}}=6.28\ \mathrm{rad/s}$. 
Four RC cars are parked as obstacles, with known positions.
The vehicle's pose is measured by a motion capture system (10 cameras, 9 markers at 100 Hz) and sent to a laptop identical to that used in simulations. 

As in simulations, the prediction horizon interval $\Delta\tau$ is fixed at 0.2\,s in real-world experiments.
For control, the cycle time $\Delta t$ is set equal to the solver runtime $t_s$ at each time $t$, since control inputs are computed online by solving the optimal control problem. 
Because $t_s$ is generally shorter than $\Delta\tau$, this approach allows $\Delta t$ to adaptively shorten, improving responsiveness to dynamic changes and modeling errors.

\subsection{Real World Experiment Results}

\begin{figure}[t]
    \begin{minipage}[t]{\linewidth}
      \centering
      \includegraphics[width=\linewidth]{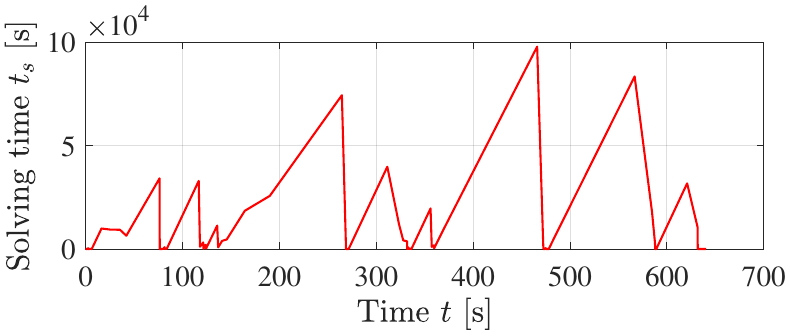}
    \end{minipage}\\
    \begin{minipage}[t]{\linewidth}
      \centering
      \includegraphics[width=\linewidth]{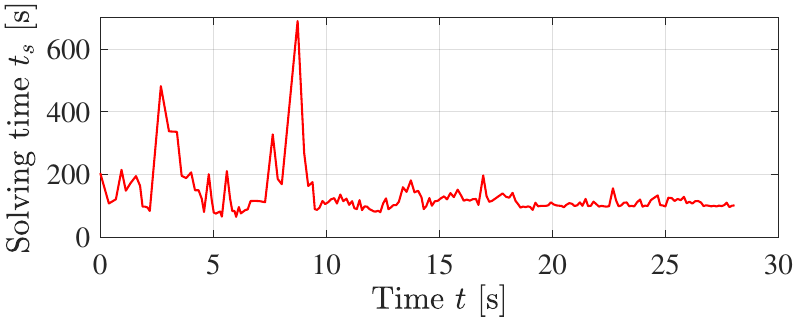}
    \end{minipage}
  \caption{Solving time for each MPC cycle in real-world experiments with SVM (top) and MSDE (bottom). Note the different time scales.}
  \label{fig:experiment_solving_time}
\end{figure}

Figure~\ref{fig:experiment_solving_time} compares the solving time per MPC cycle for SVM and MSDE in real-world experiments. 
SVM's average and worst-case solving times were 3.19\,s and 97.8\,s, respectively—often exceeding the prediction horizon $\Delta\tau=0.2\,$s—resulting in loss of control and collisions with obstacle cars. 
In contrast, MSDE's average and worst-case times were 139\,ms and 689\,ms, respectively, remaining within $\Delta\tau$. 
As a result, the RC-car smoothly parked in reverse without collisions (see Figure~\ref{fig:experiment_path}).

\begin{figure}[t]
    \begin{minipage}[t]{\linewidth}
      \centering
      \includegraphics[width=\linewidth]{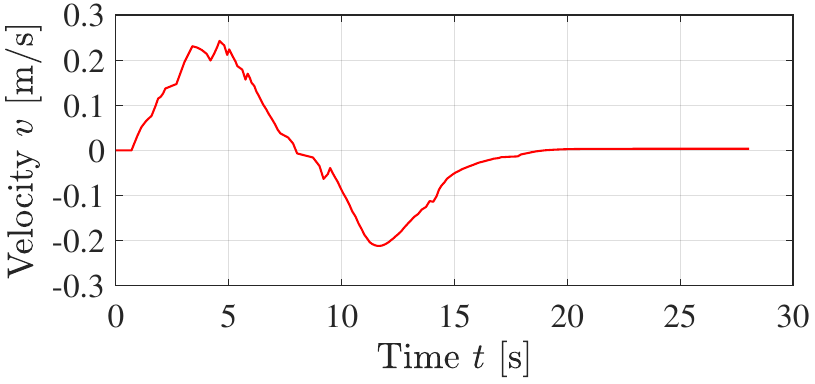}
    \end{minipage}\\
    \begin{minipage}[t]{\linewidth}
      \centering
      \includegraphics[width=\linewidth]{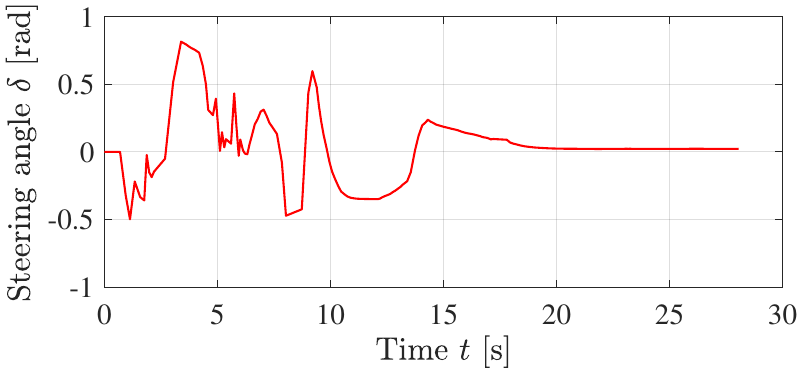}
    \end{minipage}
  \caption{Velocity and steering angle applied to the RC-car in a real-world experiment using MSDE.}
  \label{fig:experiment_control_input}
\end{figure}

Figure~\ref{fig:experiment_control_input} shows the velocity and steering angle commands under MSDE, demonstrating smooth acceleration and aggressive steering consistent with the cost weights defined in Section~\ref{sec:evaluate_function}. 
These results underscore that, for real-time vehicle control, reducing computation time is more critical than marginal gains in navigational performance.

\section{Conclusion}

This paper presents two formulations for collision-avoidance constraints within an MPC framework: SVM and MSDE. The SVM formulation is based on the separating hyperplane theorem, enforcing the existence of a separating line between two convex polygons and including a margin-maximization term to ensure convergence of the line parameters. The MSDE formulation ensures mutual non-intrusion of polygon vertices by converting disjunctive \emph{OR} constraints into conjunctive \emph{AND} constraints via a minimum function. 

Both formulations were integrated into the optimal control problem and evaluated in simulations of reverse parking, parallel parking, and obstacle avoidance in tight spaces, as well as in real-world reverse-parking experiments with an RC car. Our results demonstrate that SVM delivers superior navigation performance at the expense of greater computational cost, making it well-suited for offline motion planning. In contrast, MSDE achieves low computation times with moderate navigation performance, enabling real-time control.

Future work will extend these methods to the avoidance of dynamic obstacles and to motion planning for multiple vehicles.

\bibliographystyle{IEEEtran}
\bibliography{IEEEabrv, reference}

\end{document}